# Spherical coordinates transformation pre-processing in Deep Convolution Neural Networks for brain tumor segmentation in MRI

*Carlo Russo[a], Sidong Liu[a,b,c], Antonio Di Ieva[a,b,*]*

[a] Computational NeuroSurgery (CNS) Lab, Department of Clinical Medicine, Faculty of Medicine, Health and Human Sciences, Macquarie University, Sydney, Australia

[b] Department of Clinical Medicine, Faculty of Medicine, Health and Human Sciences, and Macquarie Neurosurgery, Macquarie University, Sydney, Australia

[c] Australian Institute of Health Innovation, Centre for Health Informatics, Macquarie University, Sydney, Australia

* Corresponding author e-mail: antonio.diieva@mq.edu.au, Computational NeuroSurgery (CNS) Lab, Macquarie University, Faculty of Medicine, Health and Human Sciences, 1st floor, 75 Talavera Rd, Macquarie Park 2109, NSW, Australia

**Abstract**

**Background and Objective:** Magnetic Resonance Imaging (MRI) is used in everyday clinical practice to assess brain tumors. Several automatic or semi-automatic segmentation algorithms have been introduced to segment brain tumors and achieve an expert-like accuracy. Deep Convolutional Neural Networks (DCNN) have recently shown very promising results, however, DCNN models are still far from achieving clinically meaningful results mainly because of the lack of generalization of the models. DCNN models need large annotated datasets to achieve good performance. Models are often optimized on the domain dataset on which they have been trained, and then fail the task when the same model is applied to different datasets from different institutions. One of the reasons is due to the lack of data standardization to adjust for different models and MR machines.

**Methods:** In this work, a 3D Spherical coordinates transform during the pre-processing phase has been hypothesized to improve DCNN models' accuracy and to allow more generalizable results even when the model is trained on small and heterogeneous datasets and translated into different domains. Indeed, the spherical coordinate system avoids several standardization issues since it works independently of resolution and imaging settings. Both Cartesian and spherical volumes were evaluated in two DCNN models with the same network structure using the BraTS 2019 dataset.

**Results:** The model trained on spherical transform pre-processed inputs resulted in superior performance over the Cartesian-input trained model on predicting gliomas' segmentation on tumor core and enhancing tumor classes (increase of 0.011 and 0.014 respectively on the validation dataset), achieving a further improvement in accuracy by merging the two models together. Furthermore, the spherical transform is not resolution-dependent and achieve same results on different input resolution.

**Conclusions:** Our model is achieving competitive dice scores and is not resolution-dependent, therefore can be applied to different datasets without re-training, thus improving segmentation accuracy and theoretically solving some transfer learning problems related to the domain shifting, at least in terms of image resolution in the datasets.

# 1. Introduction

In medical imaging, computer vision research and applications in general, Deep Learning (DL) algorithms have achieved impressive results, but only in fields where a large annotated dataset was previously available [1]. One of the major limitations of DL models is the lack of generalization when a model trained on a restricted dataset is applied to data from other institutions, introducing unpredictable results in the segmentation/classification accuracy [1]. The failure of actual models on MRI sequences is partially attributable to the lack of standardization of input images between different domains and the lack of availability of big datasets to train the DL model on a new domain. An exemplary application of a DL model is the automatic recognition and segmentation of brain tumors, such as gliomas. Since 2012, the Multimodal Brain Tumor Segmentation (BraTS) challenge is organized yearly within the International Conference on Medical Image Computing and Computer Assisted Intervention (MICCAI) (http://braintumorsegmentation.org/) [2], with the main aim of identifying the best model for automatic segmentation of MR images (MRI) of high- (HGG) and low grade- (LGG) gliomas. The segmentation task aims to automatically recognize the total area containing the tumor (called Whole Tumor, WT, region), as well as the Necrosis and Active tumor cells area (Tumor Core, TC, and Enhancing Tumor, ET, areas, the latter contained in TC area). The DL models submitted to the BraTS challenge are achieving good results when trained on a relatively small dataset, consisting of about 400 cases of HGGs and LGGs. Despite using image augmentation techniques on such a dataset, the average Dice scores of the best methods are reaching less than 90% accuracy on the WT class (average 0.88 and median 0.92), and around 80% on the ET and TC classes (average 0.78 and median 0.85 in ET, average 0.82 and median 0.91 in TC) [3]. Moreover, in spite of the heterogeneous source of the images within the BraTS dataset, there are still doubts about the translation of such models to different datasets from other institutions: indeed, some biases such as distributional shift and "black-box" decisions are still not fully addressed by current AI methodologies [4,5].

The lack of large brain tumor imaging labelled datasets and the need to overcome the heterogeneity of the datasets, as well as the need to mimic human vision and experts' interpretation, in particular neuroradiologists' expertise to identify and segment brain tumors, has triggered our current research in regards to new methods for pre-processing data representation, i.e., before feeding DL models. We here propose a new method to overcome these issues by attempting to emulate the cognitive workflow that a specialist in neuroradiology (e.g., a trained neuroradiologist or neurosurgeon) would perform for tumor recognition and segmentation.

# 2. Background and related works

*2.1. Background*

In everyday clinical practice, medical experts' eyes track scans by each single image slice, scouting for particular features in targeted anatomical regions [6,7]. During this phase, image resolution, size, and orientation of the object, for example, are not salient features of the cognitive analysis. For such a reason, DL models with strong image augmentation based on rotation and zooming usually work better than others [8]. Furthermore, radiologists in general do not look at the entire image, rather their eyes follow an apparent path and focus on one region at a time [9–14]. Our hypothesis is that DL models' accuracies can be improved by changing the representation of the image, e.g., transforming the standard Cartesian coordinates into polar-based coordinates, and the corresponding 3D spherical coordinate system for 3D volumes in the pre-processing phase. The transformation introduces a new invariant constraint compared to the classic Convolutional Neural Networks that are natively invariant to translation. Thus, the transformed object can be analyzed in a rotational/scale invariance. Furthermore, the new object will not be "position invariant" anymore, since each transformation is dependent upon the choice of an origin point. The workflow of progressive refining of the origin points during the observation process has been replicated in the steps of the new proposed method, as described below.

Polar and Spherical representation examples of segmented radiologic images of a brain tumor (a glioblastoma in the right hemisphere) are shown in Fig. 1.

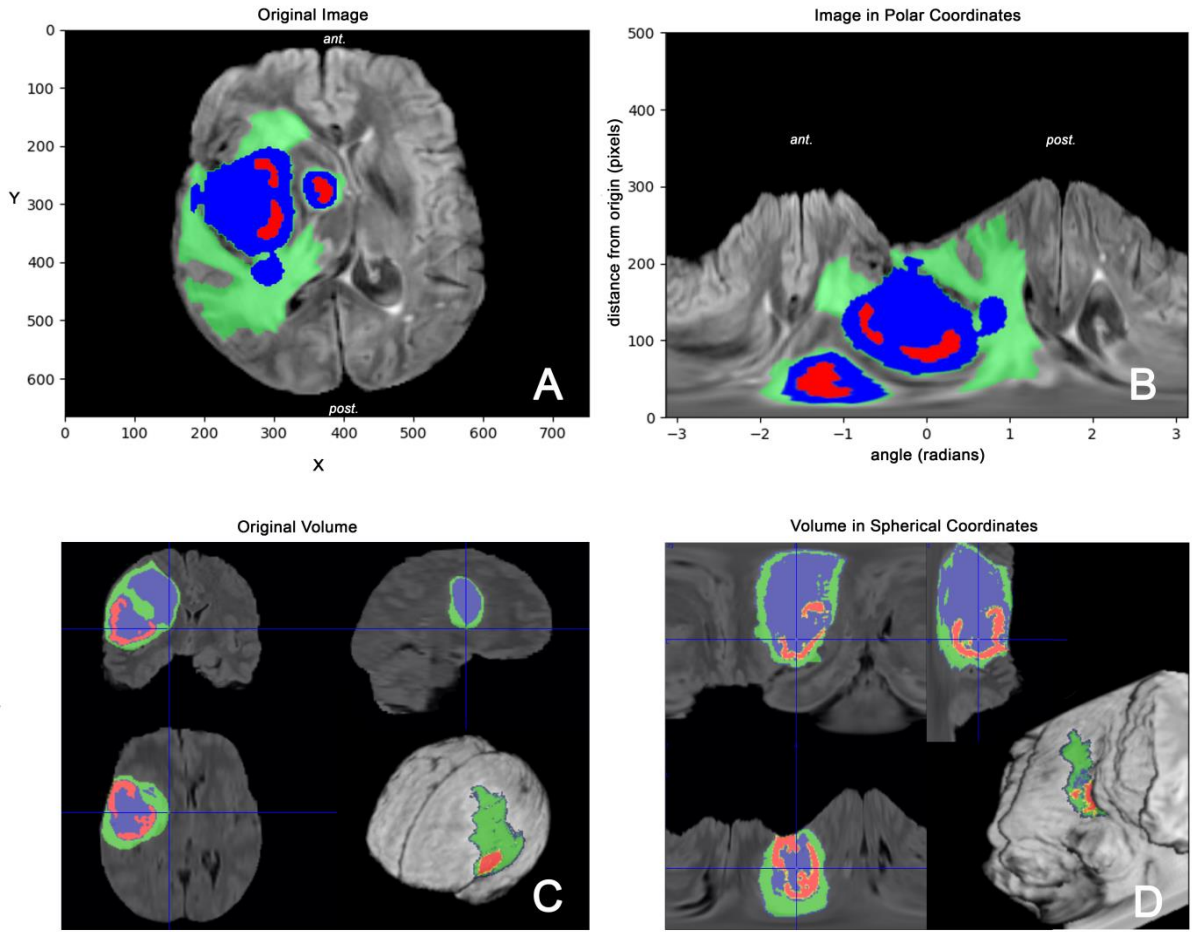

Fig. 1. Example of the representation of a radiologic image or volume in polar and spherical coordinate system. A) A single MRI image of a brain with segmented tumor areas (in green, the whole tumor, including the peritumoral edema, in red the active enhancing tumor and in blue the necrotic components within the tumor), B) The same image transformed into a polar coordinate system using the center of the image as the origin. The vertical axis represents the distance from the origin and the horizontal axis is the angle formed from each re-projected point and the origin segment with the horizontal line. C) A brain MRI volume with its 3D segmentation of the tumor, and D) the same volume transformed into a spherical coordinate system using the center of the volume as the origin. r is the distance of each re-projected point from the origin, theta and phi are the azimuth and polar angles, respectively.

A polar/spherical coordinate system pre-processing was used to feed the CNN models as it may offer the following advantages:

1) The representation of the volume in polar and spherical coordinates is invariant to rotation and scaling. This means that two volumes rotated differently have the same spherical coordinate system representation, but just shifted in the angles' planes by a distance equivalent to the rotation angle (Figs. 2C and 2D). A scaled object has the same polar representation scaled on the third direction, i.e., the distance radius, by a zooming factor equivalent to the one applied in the scaling process. This scaled representation can be avoided when considering the maximum radius to be the farthest distance from the origin to the edges of the image (e.g., the brain's surface) (Figs. 2E and 2F). Such a process enables the spherical representation of volumes to be invariant to different image resolutions within the original dataset and to different location and shapes of the tumor. Classic DL models are invariant to region shifting, therefore, by this this way they actually become invariant to scaling, rotation and resolution (at least within a specific scaling and resolution range).

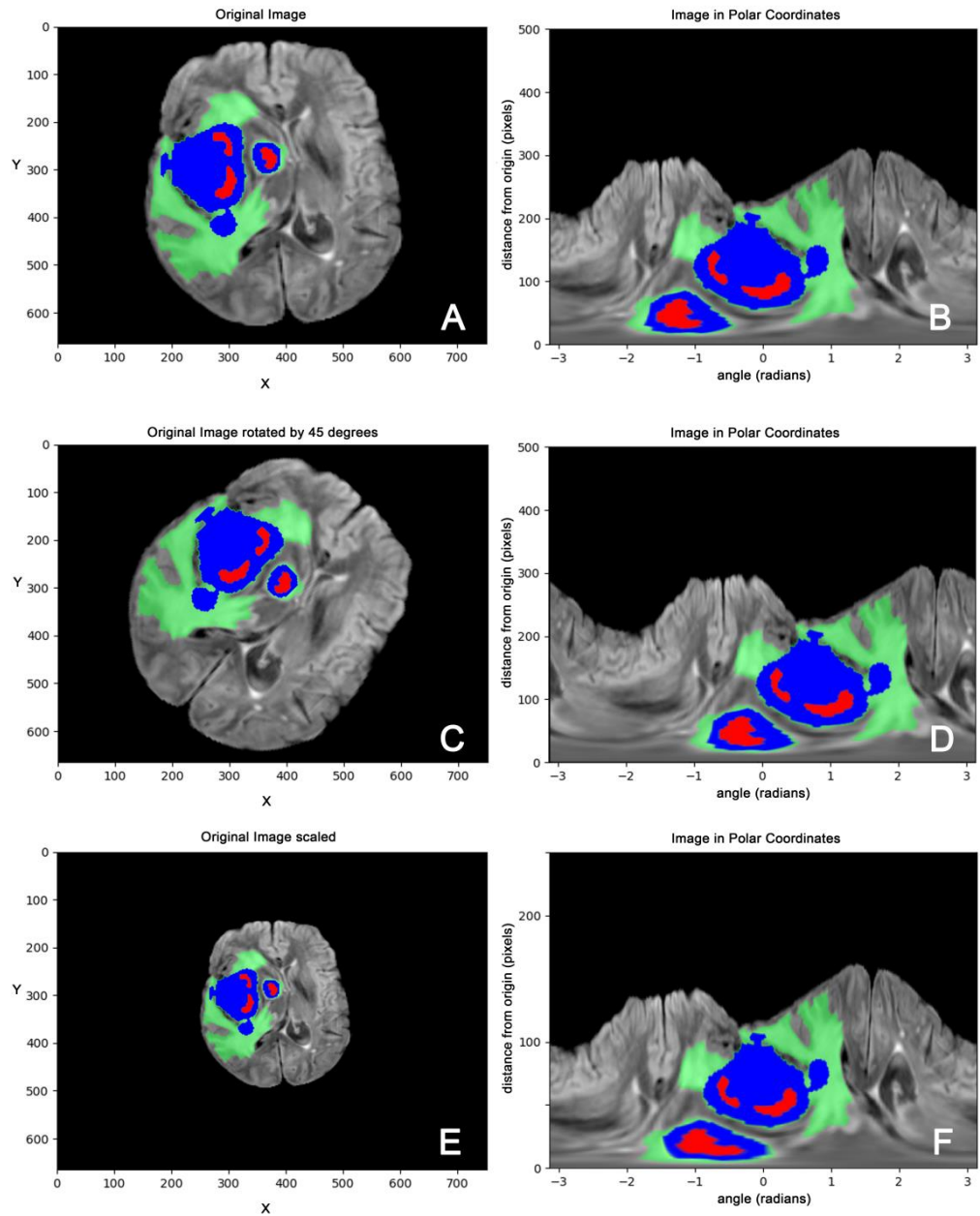

Fig. 2. Effect of rotation and scaling of the original image on its polar transformation. A) Original MRI with segmented regions of interest, B) The same image transformed into a polar coordinate system using the center of the image as the origin, C) MRI image rotated by 45 degrees, D) Polar coordinate of the rotated image, E) MRI image scaled by a factor of 2, F) Polar coordinate of the scaled image. The only effect of the rotation corresponds to a horizontal shifting of the polar transformed image. The scaling in E only scales the vertical axis and slightly deforms the bottom line since the region closest to the origin contains less details, while farthest regions contains the same amount of details.

2) The spherical system represents one region at a time. Since the spherical representation needs to be centered according to a chosen origin, the DL model will scout a single region at a time, focusing more on the centered region and minimizing the weighting of the most peripheric ones. This is also a limit of the technique but it is in fact closer to the human vision system, whereby human expert's eyes are focused on salient regions rather than on their peripheric features [6,12–15]. The ensemble of the regions will then give the fully segmented volume.

3) Spherical representation will enable the analysis of the growing surface of a centered object to be improved. The tumor biologically expands from single points growing on the border in a regular or irregular way [16]. Once the tumor segmentation is performed, this representation can be used to enable better classification of the tumor type and grade just by evaluating the irregularity of the 3D spherical surface [17], avoiding issues caused by different orientation and zooming of objects, or for any therapeutic purposes (e.g., planning of radiotherapy/radiosurgery on the targeted region, intraoperative navigation for surgery, etc.). The tumor's irregularity and complexity can be evaluated by means of radiomic parameters [18,19] and/or computational fractal-based analysis (e.g., computation of the fractal dimension), as investigated in other studies [20–27].

*2.2. Related works*

To the best of our knowledge, our study is the first attempt to use polar and spherical coordinates transform to improve image pre-processing for DL models. There have been some studies which aim to induce the rotation and scale invariance to the DCNN. For example, Harmonic Networks (H-Nets) have been proposed by Worrall et al. to improve rotation invariance [28]. In this work, DCNN has been transformed to use harmonic filters instead of regular CNN filters, thus keeping the network parameter-efficient without the necessity of including additional layers and filters for performing rotational invariance. H-Nets can be implemented in current DL frameworks with minor engineering. H-Nets are currently applicable only on 2D images, although the authors proposed to expand their work to accommodate 3D volumes [28].

Depeursinge et al. investigated the application of texture operators based on spherical harmonics wavelets achieving Local Rotational Invariance (LRI) and Directional Sensivity (DS) of radiomic features, stressing the importance of the combination of LRI and DS with popular radiomics for classifying 3D textures in particular in medical imaging where local structures of tissues occur at arbitrary rotations [29]. Cohen at al. introduced the Group Equivariant Convolutional Neural Networks (G-CNNs) to transform copies of filters inside a CNNs with shared weights to achieve rotational invariance [30]. Winkels et al. used 3D roto-translation group convolutions (G-Convs) in their 3D G-CNNs, with transformed copies of 3D filters to improve performance, sensitivity and speed of convergence for pulmonary nodule detection in volumetric CT images [31]. Andrearczyk et al. introduced LRI steerable filters and Solid Spherical Energy (SSE) as part of CNNs demonstrating superior performances compared to standard 3D G-CNNs with equivalent number of parameters [32,33]. All these works were performed by using 3D rotational invariance while trying to keep low the complexity of the resulting DCNNs and avoiding rotational image augmentation pre-processing steps. In regard to the scale invariance, it has to be emphasized that it only exists in discrete multi-step scaling of the original images. Xu et al. used Scale Invariant Convolutional Neural Network (SiCNN) as a DCNN multi-column architecture, with each column focusing on a particular scale, to automatically apply scaling transformation to the CNN layers [34]. Unlike other similar models, the interesting approach is the sharing of filter parameters within the multi-column architecture [34]. In vision research, polar representation has been shown to match human vision in a better way than other image models [35–37]. Esteves at al. presented a Polar Transformer Network (PTN) model to classify objects with rotational invariance, also extending the model to 3D objects with the use of a cylindrical coordinate system [38].

## 3. Proposed pre-processing and segmentation method

The proposed method consists in a pre-processing step to transform the input volume, followed by a repeated application of the same DCNN with different weights in order to progressively refine the segmentation. Finally, we used a post-processing step to filter excess of segmentation results by using a classical model's segmentation as well as digital filtering the final segmentation.

*3.1. The polar and spherical coordinates system*

Our method features a transformation of the volume from the standard Cartesian spatial coordinate system into the spherical coordinate system. For 2D images, changing the image coordinate system to a polar coordinate system will generate a 2D polar representation, as shown in Fig. 1A. The spherical coordinate system is equivalent to the polar transformation for 3D volumes. An example of the spherical transformation on volumes is shown in Fig. 1B. Most current MRI systems can produce 3D volumes, therefore we focused on the transformation of the spherical coordinate system and the application of 3D DCNN models. The coordinate transformation is based on the azimuth angle and polar angle of each point together with its radial distance from an origin, instead of the classical projections of the same point on the x,y,z orthogonal axes. Given an origin O having coordinates $(x_0,y_0,z_0)$ and a point $P_n$ having coordinates $(x_n,y_n,z_n)$, the new representation $T_n$ of the point $P_n$ in the spherical coordinates systems is:

$$T_n = f(P_n, O) = (r_n, \theta_n, \varphi_n)$$

where rn is the distance from $P_n$ to O, and $\theta_n$ and $\varphi_n$ are respectively the azimuth angle formed on the x,y plane by the projection of $P_n$ and x and the elevation/polar angle between z axis and the segment $OP_n$ [39] (Fig. 3). Values of rn, $\theta_n$ and $\varphi_n$ are uniformly spread on a new volume on the planes r, θ and φ, with θ ranging from π to - π, φ ranging from π/2 to – π/2 and r ranging from zero up the distance from origin to the farthest point in the original volume.

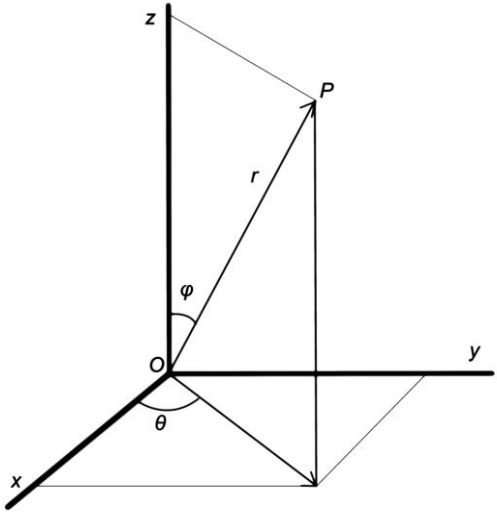

Fig. 3 Spherical coordinates of P from origin O. P has (x,y,z) coordinates in Cartesian coordinate system, as well as (r, *θ, φ*) in spherical coordinate system where r is the radial distance, *θ the* azimuthal angle, and *φ the* polar angle.

*3.2. Baseline network*

The CNN we chose to use as baseline for our method is derived from Myronenko [40], which is based on Variational Auto Encoder (VAE) U-Net with adjusted input shape and loss function according to the type of tumor to be segmented and the availability of the MRI sequences. Such a model won the BraTs 2018 competition. However, since results do not differ greatly from other simpler networks (e.g., Isensee's No New Net, second place in BraTs 2018 challenge) [8], and even if the model requires substantial resources, our choice was mainly motivated by the generalization power of Variational Autoencoders. Variational Autoencoders have been shown to be highly reliable; during the training phase the loss includes a term aimed to let the model learn a regularized Gaussian distribution of the bottleneck layers in the U-Net structure, thus enhancing regularization of the parameters of the latent space and forcing them to be continuous [41]. This kind of regularization improves the generalization of the model over unseen samples instead of matching patterns similar to those already seen during the training, thus performing better than other models on small datasets [42]. We used the Rectified Adam optimizer with a learning rate 1e-4 for the training [43]. The implementation of the network was written in Python using Keras on Tensorflow backbone, adapting a model published on Github with minor modification (https://github.com/IAmSuyogJadhav/3d-mri-brain-tumor-segmentation-using-autoencoder-regularization).

*3.3. Cascade CNN models*

The pre-processed spherical coordinates volumes are used as input on three CNN models for segmentation, in order to obtain native rotation and scaling image augmentation. An additional CNN Cartesian model is used to filter out false positive results (i.e., redundant segmentation). All of the models share the same CNN layer structure and are used in cascade. They have the same structure of the baseline CNN. Every input volume of the dataset is pre-processed using the output of the previous segmentation (except for the first pass model). The output of each segmentation is used to find the optimal origin points needed to apply the spherical transformation of the next iteration in the cascade CNNs. All of the models are trained starting with the weights obtained by the previous model training (except for the first pass model). Since the transformation uses an origin point, after a first segmentation further steps are needed to project good results backward onto the original volume. In details, the three models are the following: first pass model (origin in the center, random weights initialization), second pass model (four origins taken randomly within the first pass segmentation results, weights taken from the first pass model), third pass (four origins in the centers of the segmented object/objects obtained after the second pass, weights taken from the second pass model). The Cartesian model does not need any origin points and has a random weights' initialization. These steps refine progressively the results obtained on each iteration. Figure 4 shows an example of the process.

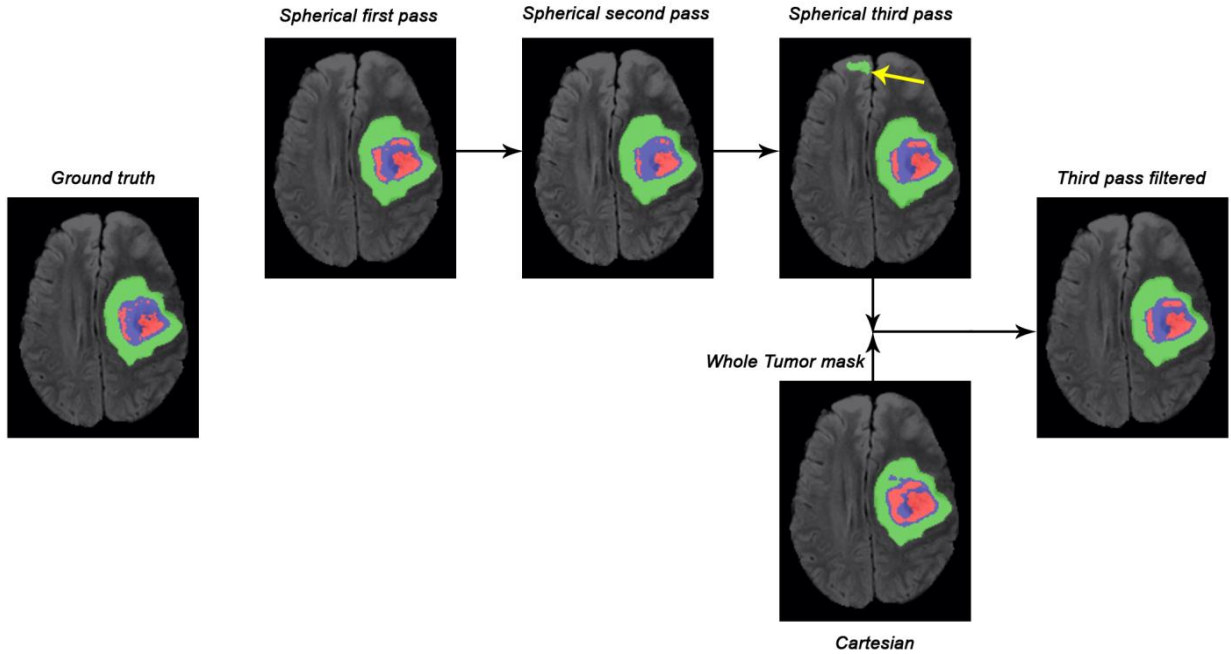

Fig. 4. An example of the application of cascade spherical CNNs and Cartesian filter to obtain final segmentation. Each image shows the segmentation result of the corresponding CNN. The final result is obtained by using the Cartesian CNN to filter the WT mask obtained on the third pass segmentation. Such a step helps to erase false positive voxel regions wrongly selected by the spherical CNNs (see segmented object indicated by the yellow arrow).

### 3.3.1. First pass.

The first pass model is used to obtain the first weights and to select the next origin point, that should be inside the initial segmentation. In the first approach on the volume segmentation, we chose the origin around the center of the MRI volume, in order to get an initial segmentation. The first pass segmentation can be also obtained by the Cartesian coordinate systems volume without any transformation, although the use of the transformation on the first pass enables the pre-trained weights to start the second pass model. To obtain a natural image augmentation, we added four further random origins inside the non-zero voxels values of the volume to the first central origin.

### 3.3.2. Second pass.

The second pass segmentation model is trained using four origins falling inside the initial segmented image. To improve the choice of the origin points for the second pass training, the first pass segmentation is filtered from isolated voxels and small objects. The choice of the origin should preferably be within larger objects, but also in other remote objects not visible enough from the first chosen origin. For such a reason, the choice of the origin points is made according to the following steps:

1. Filter of isolated voxels and small objects, apply "opening" technique (thinning and dilation) to disconnect thin connected objects and remove thin segmentations, fill small holes in the segmentation, identify individual objects from connected voxels.

2. Choose a random origin placed inside the bigger available object segmented. The origin should not be on the border of the object (thinning filter used).

3. Exclude a box having a length of 50 millimeters for each axis around the origin just considered, and repeat from step 1 until enough numbers of origin points are chosen.

4. In case not enough origin points are selected, repeat from step 1 with less limitation on the small object filtering and thinning procedure.

The values considered for the measure of the small object to be removed is 30 mm$^3$ in volume in the first iteration. Successive iterations increase this measure to 100 and 1000 mm$^3$, beginning to search into the tumor core segmentation and then expanding the search to the whole tumor segmentation until enough origin points are found.

*3.3.3. Third pass.*

A final selection of the four origin points is performed considering the center of the objects segmented in the second pass. The volume segmentation selects individual objects from connected voxels. If the origin selection is derived from a more sizeable object, greater than 50x50x50 millimeters, further origins will be taken as described in the second pass procedure, in order to not exclude remote areas and to reach the four origins needed in this pass.

*3.3.4. Cartesian CNN filter model.*

Since scale and rotation are invariant in the spherical coordinate system models, some useful information within the Cartesian spatial coordinate system can be lost. For such a reason, an additional Cartesian DCNN model has been trained on the same dataset and used to filter out false positive segmentation obtained by the spherical model.

*3.4. Image datasets and pre-processing*

Two similar resources are used to train and test our method: Medical Segmentation Decathlon (MSD) and BraTS datasets. For the training of the models we used the Medical Segmentation

Decathlon (MSD) Brain Tumors 4D Volumes Dataset [44]. This dataset includes a total of 750 4D volumes (484 Training volumes with labels and 266 Testing volumes) consisting of a selection of the BraTS 2016 and 2017 datasets. Every 4D volume contains four MR sequences of skull-stripped and co-registered images of LGGs and HGGs, namely T1, post-contrast T1, T2 and FLAIR. The Training Set contains a manually labelled ground truth. The intensity values of the source MR images are not uniform and can vary significantly amongst different sequences as well as different device setups or different machines. For such a reason, a common approach is to scale the intensities to a normal distribution centered on a value of zero and with a standard deviation of 1, only considering the non-empty voxels. In our method, the normalization was done after the transformation, in order to enhance the target region surrounding the origin point. Furthermore, brain MRI were skull-stripped before performing the transformation and intensity normalization, in order to analyze only the nervous tissue and to assure patients' anonymization.

*3.5. Image augmentation*

To augment the initial dataset, a random choice of the origin points during each training iteration was considered to be sufficient. The training phase of the first pass volume considers the origin from the center of volume as well as origins within the whole brain voxels (all of the non-empty voxels). In the second and third passes, random origins were taken inside a central region contained in the first segmentation. Even if further augmentation can be done by changing intensity of the voxels, in our cases this was already achieved since the normalization of the volume acts like an augmentation process. Indeed, for every different origin choice, a different distribution of the voxels' histogram is obtained. Thus, during the normalization of the entire volume, intensities of regions vary according to their new extension in the spherical representation.

*3.6. Label prediction*

As mentioned above, the segmentation labels in the used datasets are the three standard labels of the BraTS Challenge: 1) Whole Tumor, including all tumor and peritumoral edema, 2) Tumor Core, including necrotic, enhancing and non-enhancing tumor areas, and 3) Enhancing Tumor area. Label prediction is obtained applying the three models in the same order of the training phase. Firstly, the prediction is done with the first model choosing the origin in the center of the volume. Then, the prediction obtained is used to find the origins for the second model, and then the third model is applied. The second and third models generate more predictions of the same volume. The prediction is then summed up to reconstruct the whole segmentation. The Cartesian CNN model is then used as a filter model, applied on the original non-transformed volume to obtain another segmentation.

This last segmentation is used to intersect the spherical segmentation in order to further reduce false positive voxels. Only the Whole Tumor label of the Cartesian model is used as a filter to erase areas wrongly selected as Tumor by the spherical coordinate models, thus meaning that the new WT, TC and ET labels are obtained by the spherical segmentation excluding all the non-WT area of the Cartesian model segmentation.

*3.7. Post-processing*

As previously described, the final segmentation considers an ensemble of predictions resulting from different segmentations obtained from different models with all of their origin points and projected backwards onto the starting volume. Since the obtained segmentation still contains many false positive voxels, further post-processing steps include the removal of small segmented objects (less than 30 square millimeters in volume size), after removing thin connection between them with an "open" voxelwise operation consisting in thinning and a successive dilation of the objects' border.

*3.8. Hardware and timing*

We ran the training on a Fujitsu Celsius R970 workstation (Fujitsu Ltd. Minato, Tokyo, Japan) equipped with 2x24GB RAM NVidia P6000 GPUs (Nvidia Co., Santa Clara, California, USA). The complete training of the models required more than one month on more than 2,000 overall epochs using the 484 MSD Training Set. The segmentation of the full BraTS 2019 Validation Set of 125 patients' data required 89 minutes to complete, thus meaning that about 43 seconds per patient were needed to complete the three-pass segmentation prediction and Cartesian model filtering on our workstation.

**4. Experiment and results**

*4.1. The training phase*

The three CNN models were trained in cascade on the MSD Training Dataset. We used all of the 484 MSD Training Set for the training phase as follows: 400 input volumes were used for the training and 84 input volumes for the validation. The input volume generator applies the spherical transformation on every iteration on both the input volume and the ground truth choosing the random origins according to the steps of the method described above for every one of the first, second and third pass models. So, after the first pass model was completely trained, we used the obtained segmentation to get the origin space for the second pass model. After obtaining the second pass model segmentation, the third pass model was fed with the origins located near the centers of the segmented objects derived from the second pass predictions.

*4.2. Evaluation*

To compare our results to the state-of-the-art on automatic tumor segmentation achieved so far, we chose to apply it on the BraTS 2019 training and validation datasets and to use the online evaluation platform from the CBICA Image Processing Portal (A web platform for imaging analysis, Centre for Biomedical Image Computing and Analytics, University of Pennsylvania, USA: https://ipp.cbica.upenn.edu/). [3,45,46]. While the training process uses a Soft Dice Score-based loss (along with the Least Square Errors loss of the VAE reconstruction branch and Kullback-Leibler divergence evaluated on the latent space, as described by Myronenko) [40], we uploaded the final segmentations to the official BraTS 2019 server for evaluation of per class dice, sensitivity, specificity and Hausdorff distances directly on the BraTS 2019 Training and Validation datasets. We did not use BraTs testing datasets since the platform only allows to get official evaluation for training and validation datasets. Furthermore BraTs 2019 testing dataset is not available to those not participating to the competition. The advantage of also evaluating the BraTS 2019 Training dataset is the differentiations of the scores on LGG and HGG groups, while the Validation Set does not distinguish the two glioma groups.

*4.3. Comparison with baseline*

The Cartesian model was used as a baseline for our comparison. The spherical model was applied on the same dataset. Then the postprocessing phase was used in the final step to remove false positive voxels. As shown in Table 1, the Spherical coordinates model alone did not improve much the overall results obtained with only the baseline model.

Table 1. Dice scores, sensitivity, specificity and Haussdorff 95 of each segmentation label predicted by the models on the BraTS 2019 training and validation datasets.

| | Brats 2019 Training dataset | | | | | | | | | | | |
|---|---|---|---|---|---|---|---|---|---|---|---|---|
| | Dice Score | | | Sensitivity | | | Specificity | | | Hausdorff 95 | | |
| | WT | TC | ET | WT | TC | ET | WT | TC | ET | WT | TC | ET |
| Spherical model | 0.882 | 0.803 | 0.717 | **0.940** | **0.888** | **0.823** | 0.987 | 0.994 | 0.997 | 7.351 | 7.220 | 5.691 |
| Cartesian (baseline) model | 0.890 | 0.792 | 0.724 | 0.910 | 0.834 | 0.819 | 0.993 | **0.996** | **0.998** | 7.207 | 8.630 | 5.753 |
| Intersection and postprocess | **0.902** | **0.805** | **0.755** | 0.893 | 0.873 | 0.816 | **0.995** | 0.995 | 0.997 | **4.755** | **5.903** | **4.081** |
| | Brats 2019 Validation dataset | | | | | | | | | | | |
| | Dice Score | | | Sensitivity | | | Specificity | | | Hausdorff 95 | | |
| | WT | TC | ET | WT | TC | ET | WT | TC | ET | WT | TC | ET |
| Spherical model | 0.847 | 0.762 | 0.704 | **0.941** | **0.836** | **0.792** | 0.985 | 0.994 | 0.997 | 11.550 | 12.220 | 8.496 |
| Cartesian (baseline) model | 0.873 | 0.749 | 0.690 | 0.915 | 0.764 | 0.749 | 0.992 | **0.997** | **0.998** | 10.250 | 12.790 | 7.925 |
| Intersection and postprocess | **0.879** | **0.764** | **0.737** | 0.899 | 0.819 | 0.786 | **0.993** | 0.995 | 0.997 | **8.655** | **10.840** | **6.811** |

On the training dataset, there was a slight overall improvement in the Tumor Core segmentation score (0.011) whilst WT and ET segmentation scored a similar dice value (difference inferior to 0.008). On validation dataset, a similar trend was observed for TC, while WT results were worse than on the training dataset and ET results improved slightly. However, as shown in fig. 4, we observed that segmentation results of the Spherical coordinate system were actually selecting the main tumor objects better, but the overall score was worsening for the inclusion of many false positive areas selecting normal brain structure having the same characteristics of tumor areas. This is also confirmed by the lower specificity score of the spherical model, as shown in Table 1. These areas were not included in the Whole Tumor areas of the Cartesian model. The final score was then obtained by using the spherical coordinate model filtering the non-WT areas of the segmentation obtained by the Cartesian model and then applying the post-processing to remove small isolated groups of false positive voxels. After this filtering, the final dice score as well as the specificity improved in all the three classes. The computation of another measure for segmentation accuracy used in the BraTS challenge, the Hausdorff 95, also demonstrated the improvement of the segmentation.

We also evaluated the difference between the score of the two LGG and HGG groups in the BraTS Training dataset, and we found the overall TC and ET segmentation were better in the HGG dataset compared to the LGG one, while WT was similar (see Table 2).

Table 2. Dice score per class.

|  | Full training dataset | | | HGG only | | | LGG only | | |
| --- | --- | --- | --- | --- | --- | --- | --- | --- | --- |
|  | WT | TC | ET | WT | TC | ET | WT | TC | ET |
| Spherical model | 0.882 | 0.803 | 0.717 | 0.885 | 0.854 | 0.804 | 0.868 | **0.627** | 0.417 |
| Cartesian model | 0.890 | 0.792 | 0.724 | 0.893 | 0.847 | **0.821** | 0.879 | 0.601 | 0.394 |
| Intersection and postprocess | **0.902** | **0.805** | **0.755** | **0.907** | **0.858** | 0.802 | **0.884** | 0.625 | **0.592** |

*4.4. Application of the model on different resolution volumes*

Since the spherical transformation uses natively scaled and rotationally invariant coordinate systems, its application to a different resolution dataset is trivial. However, to validate this, we rotated and zoomed the original dataset randomly and then applied the Spherical transformation, obtaining similar resulting volumes, as shown in Fig. 2. The transformation also introduced different window and level intensity values to the voxels after the normalization step. The BraTS Training and Validation models had previously been transformed and fed into the new method, obtaining similar segmentation (see Fig. 2). It is worthy to mention that to find the WT label used to filter spherical segmentation, the Cartesian model was applied to the isotropic transformation of the volume, because this model still has the limitation of the isotropic resolution of the dataset. However, this WT label was only used to filter out the excess of segmentation and, whenever not available (for example, when the volume included only part of the brain), it was always possible to avoid this step.

## 5. Discussion

Spherical coordinate transformation is introducing an extreme augmentation. Each input volume of the brain tumor dataset is modified at every iteration during the training of the model for each origin point randomly chosen. With this augmentation and its invariance to rotation and scaling, the adoption of spherical coordinate system helps to apply the method to datasets coming from environments that use different resolution and settings. The extreme augmentation of the spherical model strengthens the robustness of the DCNN, however, it can occasionally lose information in regards to normal brain structures. Thus, variable accuracies are obtained in both spherical and Cartesian models when applied in distinct regions. We have shown for the first time that the combination of the Cartesian and spherical models together improves the results when compared to the single models alone. Compared to other similar works focused on rotational and scale invariance, our method does not need any change of any Deep Learning frameworks, since it involves only a pre-processing step to transform the coordinate system before using any CNN architecture. Like H-Nets this method is invariant to any rotation, not only to 90-degree rotations. This method achieves scale invariance like the SiCNN without the need to replicate columns along CNN layers and copying filters, since the same filters already work on multiscale transformation. The Polar Transformer Network and G-CNNs use a similar approach, but incorporates the transformation into the CNN architectures. PTN also works with 3D objects using a cylindrical coordinate system instead of spherical coordinates. Our spherical coordinate systems also give rotational invariance in the elevation angle while cylindrical coordinate system only enables a polar angle invariance on 3D objects. 3D G-CNNs work with 3D objects as well, but do not provide scale invariance. Our new method did not score within the best BraTS 2019 competition models. However, more importantly our method achieved competitive results compared to other DCNN current methods and can be used on different environments and datasets as it is not limited to a specific volume resolution. Furthermore, the neurosurgeon (ADI) involved in the study as well as neuroradiologists who were blinded to the results manually checked the volumes and found that our method corrected some wrong ground truth segmentation on the BraTS' Training Dataset (see Fig 5). By not having the ground truth segmentation of the validation dataset, it was not possible to check the accuracy of the validation ground truth on cases when we obtained worse results.

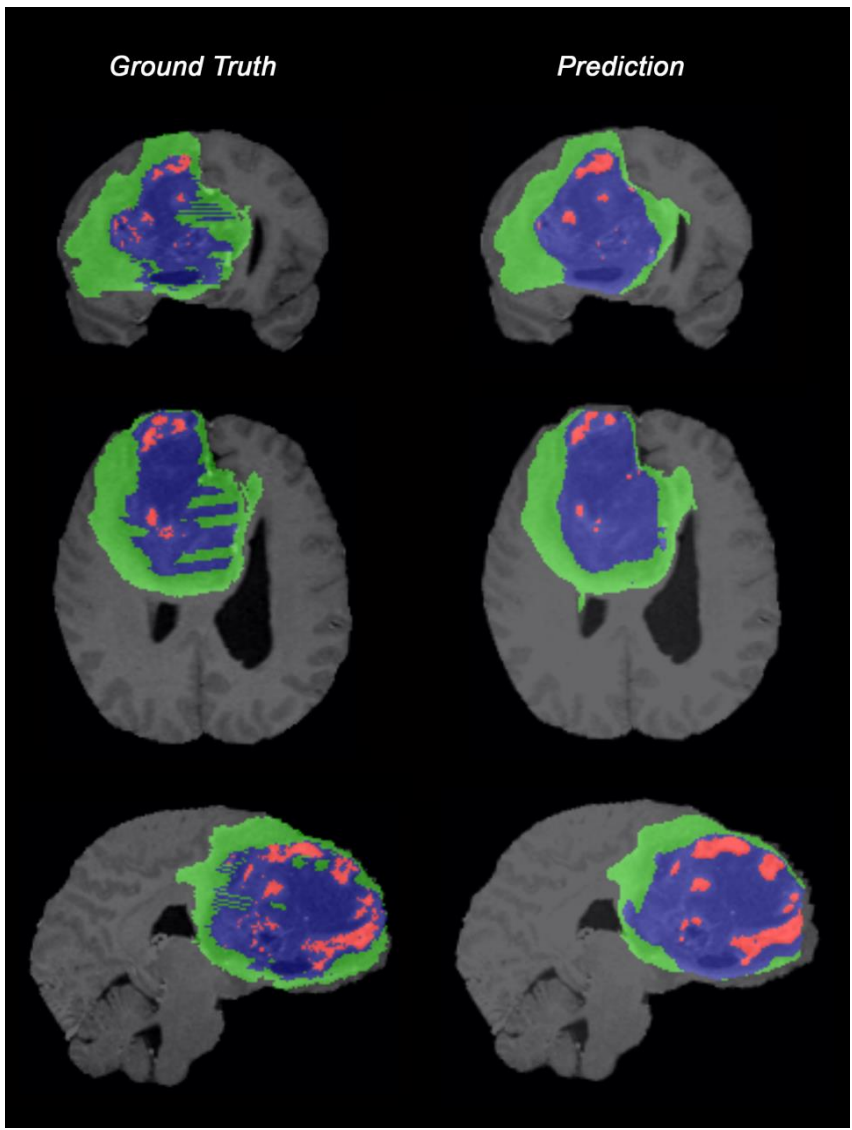

Fig. 5. On the left, an example of a ground truth found in BraTS 2019 dataset, containing a questionable segmentation. The segmentation predicted by the model (on the right) is smoother and appears a more realistic match.

**6. Conclusions**

We here introduce a novel approach aimed to change the way DL models are fed with the goal of improving reliability and avoiding limitations of resolution and dataset domain standardization.

This new approach includes a native image augmentation related to the spherical transformation technique applied during the pre-processing of input data. This transformation mimics human expert perception, focusing on details and single regions during the segmentation process. Since the spherical coordinate system models are invariant to scale and rotation, as a drawback they may occasionally lose some useful information of the Cartesian spatial coordinate system. For such a reason, the best performance is achieved when a Cartesian coordinate model is preliminarily used to filter out the non-tumor areas. Our results improved upon the baseline Cartesian model. Our model is achieving competitive dice scores and is not resolution-dependent, therefore can be applied to different datasets without re-training, thus theoretically solving some transfer learning problems related to the domain shifting, at least in terms of image resolution in the datasets. Although only a slight increase in accuracy has been found so far and further studies are required to prove the better performance of the new method to solve the shifting domain problem, experiments based on rescaled and rotated volumes of the BraTS dataset proved the robustness of this method on different settings of the scan parameters of position, resolution and intensity values, thus the method is expected to work better and in real world scenarios. Future studies will involve a collection and manual segmentation of new data using original resolution volumes from our and other institutions, as well as the application of the method to further available datasets.

**Funds**

Prof. Antonio Di Ieva received the 2019 John Mitchell Crouch Fellowship from the Royal Australasian College of Surgeons (RACS), which, along to Macquarie University co-funding, supported the opening of the Computational NeuroSurgery (CNS) Lab at Macquarie University, Sydney, Australia, where this work was performed. Moreover, he is supported by an Australian Research Council (ARC) Future Fellowship (2019-2023, FT190100623). Dr. Sidong Liu is supported by an Australian National Health and Medical Research Council grant (NHMRC Early Career Fellowship 1160760). None of these agencies were directly involved in this research.

**Acknowledgments**

Special thanks to Jennilee Davidson for proofreading of the paper.

**CRediT Author Statement**

*Carlo Russo:* Conceptualization, Methodology, Software, Validation, Formal Analysis, Investigation, Data curation, Writing – Original Draft, Writing – Review & Editing, Visualization

*Sidong Liu:* Validation, Investigation, Data curation, Writing – Review & Editing

*Antonio Di Ieva:* Conceptualization, Investigation, Resources, Writing – Review & Editing, Visualization, Supervision, Project administration, Funding acquisition.